\documentclass{article}

\usepackage{microtype}
\usepackage{graphicx}
\usepackage{subfigure}
\usepackage{booktabs} 

\usepackage{hyperref}

\usepackage[accepted]{icml2023}

\usepackage{amsmath}
\usepackage{amssymb}
\usepackage{mathtools}
\usepackage{amsthm}

\usepackage[capitalize,noabbrev]{cleveref}

\theoremstyle{plain}

\theoremstyle{definition}

\theoremstyle{remark}

\usepackage[textsize=tiny]{todonotes}

\DeclareMathOperator*{\argmin}{arg\,min}

\begin{document}

\twocolumn[
\icmltitle{Context-Aware Self-Adaptation for Domain Generalization}



\icmlsetsymbol{equal}{*}

\begin{icmlauthorlist}
\icmlauthor{Hao Yan}{carleton}
\icmlauthor{Yuhong Guo}{carleton,amii}
\end{icmlauthorlist}

\icmlaffiliation{carleton}{School of Computer Science, Carleton University, Ottawa, Canada}
\icmlaffiliation{amii}{Canada CIFAR AI Chair, Amii, Canada}

\icmlcorrespondingauthor{Hao Yan}{haoyan6@cmail.carleton.ca}
\icmlcorrespondingauthor{Yuhong Guo}{yuhong.guo@carleton.ca}

\icmlkeywords{Domain generalization}

\vskip 0.3in
]



\printAffiliationsAndNotice{}  

\begin{abstract}
Domain generalization aims at developing suitable learning algorithms in source training domains such that the model learned can generalize well on a different unseen testing domain. 
We present a novel two-stage approach called Context-Aware Self-Adaptation (CASA) for domain generalization. CASA simulates an approximate meta-generalization scenario and incorporates a self-adaptation module to adjust pre-trained meta-source models to the meta-target domains while maintaining their predictive capability on the meta-source domains. 
The core concept of self-adaptation involves leveraging contextual information, such as the mean of mini-batch features, as domain knowledge to automatically adapt a model trained in the first stage to new contexts in the second stage.
Lastly, we utilize an ensemble of multiple meta-source models to perform inference on the testing domain.
Experimental results demonstrate that our proposed method achieves state-of-the-art performance on standard benchmarks.
\end{abstract}

\section{Introduction}

Deep models generalize well if the training and testing data share the same data distribution. However, when the testing data distribution deviates from the training data, model fine-tuning becomes necessary, requiring additional computational resources that may not be available in resource-constrained applications. 
Moreover, collecting and labeling additional training data can be time-consuming and impractical for real-time inference scenarios.
To address this challenge, domain generalization (DG) methods have emerged as a promising solution. DG aims to train models on training domains with the expectation that they will generalize well to unseen testing domains. Existing DG methods employ various approaches, such as enriching the source data distribution through data augmentation techniques \cite{volpi2018generalizing, zhou2020learning} or learning domain-invariant models \cite{li2018deep, mahajan2021domain}. Some methods utilize meta-learning to mimic the generalization process, enabling the learning of models that exhibit generalization capabilities \cite{li2018learning}. 
Additionally, the ensemble of multiple source models has been explored as a means to enhance DG performance \cite{arpit2021ensemble}.
The primary challenge in domain generalization lies in the lack of access to the testing domain and the limited quantity of source domains available.

In light of these challenges, we propose a novel two-stage method, named Context-Aware Self-Adaptation (CASA) for domain generalization. 
CASA simulates an approximate meta-generalization scenario 
and designs a simple but fresh self-adaptation module to adjust pre-trained meta-source models to the meta-target domains 
while maintaining their predictive capability on the meta-source domains. 
This is realized through a two-stage training process: 
A regular model is trained in the first stage on the meta-source domain, 
while a self-adaptation module is added in the second stage that 
leverages contextual information, such as the mean of mini-batch features, as domain knowledge to automatically 
adapt the  model trained in the first stage to new contexts in the second stage.
To ensure that the adapted meta-target feature vectors remain in the same vector space as the original vectors, 
which plays a key factor for maintaining generalizability on the original training data,
we propose a context-aware feature-wise linear modulation (CaFiLM) mechanism
for self-adaptation.
This involves learning dimension-specific weights and biases from the feature vector and the context information.
These learned weights and biases are then used to linearly modulate the meta-target features. 
Finally, we perform inference on the testing domain by utilizing an ensemble of multiple adapted meta-source models.
To demonstrate the effectiveness of our proposed method, we evaluate it on widely used domain generalization benchmarks.
The experimental results show that the proposed CASA method surpasses the state-of-the-art approaches on small-scale datasets and achieves comparable results on large-scale datasets, producing the best overall performance among a great number of competitors.

\section{Related Work}

Domain generalization (DG) aims at training a model based on the source training data and expects the model to generalize well on the unseen testing testing data. Existing DG methods can be broadly divided into three categories \cite{wang2022generalizing}: data augmentation, domain-invariant representation learning and learning strategy design. 
Data augmentation based methods try to enrich source data distribution with the expectation of covering testing data distribution as much as possible. Typical data augmentation techniques for DG include MixUp \cite{wang2020heterogeneous}, adversarial data augmentation \cite{volpi2018generalizing} and data generation \cite{zhou2020learning}. 
Domain-invariant DG methods
aim at learning domain-invariant features. 
Various domain-invariant representation learning techniques have been used for DG,
including
adversarial domain alignment \cite{li2018deep},
which were widely used to 
reduce the cross-domain feature discrepancy. 
In addition, several criteria are introduced to measure the domain discrepancy and minimized to learn domain-invariant representations, including Maximum Mean Discrepancy (MMD) \cite{li2018domain}, Kullback–Leibler (KL) divergence \cite{li2020domain} and contrastive loss \cite{mahajan2021domain}. 
Adopting learning strategies for training generalizable models is another line of research for DG.
In particular, meta-learning is introduced to mimic the generalization procedure by splitting meta-train and meta-test domains from the multiple source domains \cite{li2018learning},
with a comprehensive coverage 
in the survey papers \cite{wang2022generalizing, zhou2022domain}.

Recently, a unified DG framework DomainBed \cite{gulrajani2020search} has been proposed and 
various previous methods are evaluated under the same fair circumstance. It shows that the simplest baseline Empirical Risk Minimization (ERM) outperforms most of the previous DG methods under the fair comparison. 
This leads to a set of new methods that tackle DG from different perspectives. 
SWAD \cite{cha2021swad} seeks flat minima of ERM through stochastic weight averaging. mDSDI \cite{bui2021exploiting} jointly learns domain-invariant and domain-specific features through the meta-learning framework. DNA \cite{chu2022dna} derives a PAC-Bayes upper bound on the target square-root risk and minimizes the bound approximately. EoA \cite{arpit2021ensemble} boosts the DG performance by ensembling independently trained moving average models with different hyper-parameters and random seeds. AGFA \cite{kim2023domain} creates synthetic data by generating Fourier amplitude images and utilizes domain adaptation methods to train the classifier. 

Our proposed CASA method in this work differs from these existing methods in several key aspects: 
we approximate a model-upgrade based two-stage meta-generalization scenario with the combination of limited source domains;
the novel design of the self-adaptation module promotes generalizability across both meta-source and meta-target domains;
and the characterisic of contex-aware adaptation brings forth the automatic adaptation ability.

The self-adaptation capacity introduced by our proposed model is also fundamentally different 
from the source-free domain adaptation techniques.
Source-free domain adaptation aims at adapting the pre-trained source model to the target domain with access to the unlabeled target data
\cite{liang2020we,lee2022confidence, yan2022dual,yan2021source}, 
which entirely focuses on the given target domain. 
In contrast, our method propels the capacity for broader generalization by accounting for unknown target domains.
\begin{figure*}[t]
    \centering
    \includegraphics[width=0.8\textwidth]{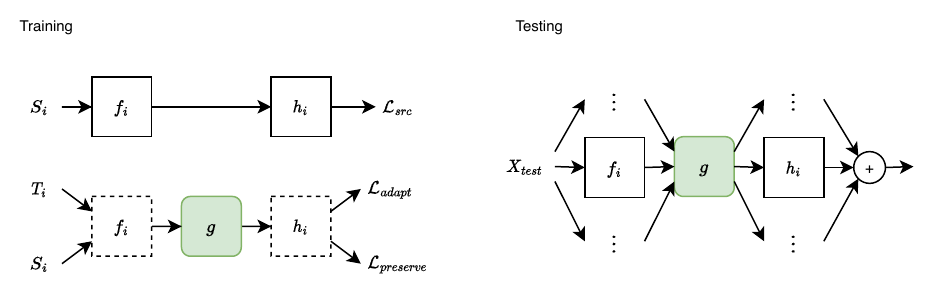}
    \caption{Diagram of the proposed two-stages Context-Aware Self-Adaptation method for domain generalization. Left upper: One model is trained for each meta-source domain. Left lower: Context-Aware Self-Adaptation module is trained to adapt the pre-trained meta-source model to the meta-target domain while preserving the prediction ability on the meta-source domain. Right: Ensemble of the adapted meta-source models are used for testing.}
    \label{diagram}
\end{figure*}

\section{Proposed Method}

This paper addresses the following domain generalization (DG) setting. Given the labeled data from $d$ 
source 
training domains $D=\{D_j\}_{j=1}^{d}$, the goal is to train a model that generalizes well on an unseen testing domain. To tackle this problem, we propose a novel two-stages method called Context-Aware Self-Adaptation (CASA). 
CASA simulates an approximate meta-generalization scenario by contructing a set of meta-source and meta-arget domain pairs
from the training domains. 
In the first stage, each meta-source domain is used to train a meta-source model. 
In the second stage, we introduce a context-aware self-adaption module between the pre-trained meta-source feature extractor and classifier. 
This module is designed to adapt the pre-trained model to the meta-target domain while preserving the prediction ability on the meta-source domain. Finally, we employ an ensemble of multiple adapted meta-source models for inference on the testing domain. The proposed method is illustrated in Figure~\ref{diagram}, and will be elaborated below.

\subsection{Two-Stage Generalizable Learning}
In view of the main obstacle of domain generalization---the test domain is unknown and not accessible during the training stage,
we propose to simulate a meta-generalization environment for learning generalizable mechanisms. 
In this meta-generalization environment, we simulate a set of meta-source and meta-target domain pairs, 
and design a two-stage training procedure to allow 
automatic adaptation of pre-trained models in the second stage with a self-adaptation module.

Specifically, 
given the data from $d$ training domains $D=\{D_j\}_{j=1}^{d}$, we pick one or multiple domains to form the meta-source domain $S_i$
and keep the rest as the corresponding meta-target domain $T_i$:
\begin{equation}
    S_i = \cup_{j=1}^d \boldsymbol{\delta}_i(j) D_j,\quad T_i = D\backslash S_i.
\end{equation}
where $\boldsymbol{\delta}_i(j)$ is $1$ if $D_j$ is picked to form $S_i$ and $0$ otherwise. 
Repeating this process, we can construct a set of meta-source and meta-target domain pairs:
$\mathcal{T} = \{(S_i, T_i)\}$.

For the two-stage training procedure, 
we train a regular meta-source model for each meta-source domain $S_i$ in the first stage. 
Considering the classification task, we denote the classification model as the concatenation 
of a feature extractor $f_i(\cdot;\boldsymbol{\theta}_{f_i})$ and a classifier $h_i(\cdot;\boldsymbol{\theta}_{h_i})$. 
The classification model is thus denoted as $h_i\circ f_i(\cdot;\boldsymbol{\theta}_{f_i},\boldsymbol{\theta}_{h_i})$. For each meta-source domain $S_i$, we train one model $h_i\circ f_i$ with empirical risk minimization, i.e.
\begin{equation}
\begin{split}
    \boldsymbol{\theta}_{f_i}^{*}, \boldsymbol{\theta}_{h_i}^{*} &= \argmin_{\boldsymbol{\theta}_{f_i}, \boldsymbol{\theta}_{h_i}} \mathcal{L}_{src},\\
    \mathcal{L}_{src} &= \mathbb{E}_{(x,y)\in S_i} \ell_{ce}(h_i\circ f_i(x;\boldsymbol{\theta}_{f_i},\boldsymbol{\theta}_{h_i}),y)
\end{split}
\end{equation}
where $\boldsymbol{\theta}_{f_i}^{*}, \boldsymbol{\theta}_{h_i}^{*}$ denotes the optimal parameters of the feature extractor and classifier,
$(x,y)$ denotes a pair of instance and label sampled from the training data of the meta-source domain $S_i$ and $\ell_{ce}$ denotes the cross-entropy loss.

Given the pre-trained meta-source model $h_i\circ f_i$ and the meta-target domain $T_i$,
a typical strategy to adapt to the target domain is to deploy the fine-tuning technique. 
However, the fine-tuning process only seeks to adapt to the specific target domain,
providing no generalization capacity in light of unknown test domains. 
In order to enable generalization to unseen target domains, we need to learn an adaptation mechanism 
that captures the intrinsic nature of adapting the pre-trained meta-source models to different meta-target domains. 
To this end, we plan to deploy a feature representation adaptation module 
that can automatically adapt the features extracted from the pre-trained meta-source feature extractor
in a context-aware manner,
allowing the pre-trained meta-source classifier to work effectively with the adapted features from the meta-target domain. 
Simultaneously, in order to ensure genuine context-aware adaptation, it is reasonable to expect that the upgraded model will continue to perform well in the original meta-source domain.

Following this design scheme,
we introduce a task-invariant adaptation module denoted as $g$ in the second stage, 
which is inserted between the feature extractor $f_i$ and classifier $h_i$,
and upgrades each pre-trained model $h_i\circ f_i$ to a combination
model $h_i\circ g\circ f_i$, as shown in Figure~\ref{diagram}. 
This module $g$ is shared across the set of meta adaptation tasks $\mathcal{T}$. 
The purpose of this adaptation module $g$ is to learn how to adapt 
the features extracted by $f_i$ 
to work well with a fixed classifier $h_i$,
given the current task context $\mathcal{C}(\cdot)$.
The objective for training the adaptation module $g$ on the set of meta-target domains from $\mathcal{T}$ is defined as follow,
\begin{equation}
\begin{split}
    \min_{\boldsymbol{\theta}_{g}} 
    &\mathcal{L}_{adapt} =\mathbb{E}_{T_i\in \mathcal{T}} \mathbb{E}_{(x,y)\in T_i} \\ &\ell_{ce}(h_i\circ g_{\mathcal{C}(T_i)}\circ f_i(x;\boldsymbol{\theta}_{g},\boldsymbol{\theta}_{f_i}^{*},\boldsymbol{\theta}_{h_i}^{*}),y)
\end{split}
\end{equation}
where $(x,y)$ is a pair of instance and label sampled from the meta-target domain $T_i$,
and $\mathcal{C}(T_i)$ denotes the context information for domain $T_i$.
The model parameters of the pre-trained meta-source feature extractor and classifier remain fixed and the optimization is focused solely on the parameters of the adaptation module $g$.

However, when adapting pre-trained meta-source models 
using the objective above, 
there is a risk that the additional adaptation module and the 
adaptation process might compromise the prediction ability on the original meta-source domains. 
This means that the best-performing adaptation module may not necessarily be the one that retains strong performance on all the training domains. 
To address this issue, it is crucial to introduce an additional objective that specifically preserves the prediction ability on the original meta-source domain as follow:
\begin{equation}
\begin{split}
    \min_{\boldsymbol{\theta}_{g}} 
    &\mathcal{L}_{preserve} =\mathbb{E}_{S_i\in \mathcal{T}} \mathbb{E}_{(x,y)\in S_i} \\
    &\ell_{ce}(h_i\circ g_{\mathcal{C}(S_i)}\circ f_i(x;\boldsymbol{\theta}_{g},\boldsymbol{\theta}_{f_i}^{*},\boldsymbol{\theta}_{h_i}^{*}),y)
\end{split}
\end{equation}
where $(x,y)$ is a pair of instance and label sampled from the training data of the meta-source domain $S_i$, 
and $\mathcal{C}(S_i)$ denotes the context information for domain $S_i$.
This objective is also essential to induce genuine context-aware adaptation
that can adjust the prediction based on the current given context.
The overall training objective in the second stage for learning the adaptation module is
\begin{equation}
    \min_{\boldsymbol{\theta}_{g}} 
    \mathcal{L}_{adapt} + \lambda \mathcal{L}_{preserve}
\end{equation}
where $\lambda$ is the trade-off hyper-parameter.
We expect such a two-stage learning procedure can induce a context-aware generalizable model
that can readily adapt to new contexts.

\subsection{Context-Aware Self-Adaptation Module}

The design of the adaptation module $g$ 
involves determining the input information,
in particular, the form of context representation, and the structure of the module itself. 
One straightforward approach is to use a multi-layer perceptron (MLP) model as a universal approximation function to process each feature vector corresponding to an instance. 
However, relying solely on instance-wise processing does not guarantee generalization to unseen testing domains, 
since 
an identical instance has the potential to be perceived or classified differently within varying domain contexts.
To address this challenge, 
the module should possess the capability to swiftly decode the domain context in the inference process
and dynamically adapt to the provided data.
This places significant demands on the choice of the domain context information and the design of the adaptation module.

\subsubsection{Choice of Context Information}
One of the most common and easily acquired local context information is the mini-batch feature mean. 
Deep models are typically trained using randomly sampled mini-batches, which are considered to represent the global data distribution. 
Similarly, during testing, deep models often process data in mini-batches to achieve faster inference speeds. 
This mini-batch mechanism conveniently allows us to obtain the context information in the form of mini-batch feature mean,
which can be considered as an approximation of the domain feature mean during the training and inference processes. 
By incorporating the mini-batch feature mean as 
an additional input to $g$ during training, 
the self-adaptation module can learn to interpret useful context information from it to facilitate the instance feature adaptation.
As a result, the input information to the self-adaptation module includes both the instance feature vector and the mini-batch feature mean;
and the $g$ function can be expressed as:
\begin{equation}
	g_{\mathcal{C}(A_i\in\mathcal{T})} = 
	g_{\mathcal{C}(x)} = 
	g(f_i(x), \mathbb{E}_{x\in X_b} f_i(x);\boldsymbol{\theta}_g)
\end{equation}
where $A_i$ denotes either a meta-source domain $S_i$ or a meta-target domain $T_i$,
and $\mathcal{C}(x)$ denotes the local context information around $x$
that can be computed as the 
mini-batch data $X_b$ around instance $x$ in the corresponding domain $A_i$. 

Incorporating the mini-batch feature mean as context information does not introduce complex processing during the training procedure. Unlike requiring the frequent calculation of a global domain feature mean at each iteration step, using the mini-batch feature mean is a computationally efficient approach. Furthermore, it does not alter the inference procedure as there is no need to calculate a global domain feature mean prior to inference.
It is important to note that the context information provided by the mini-batch feature mean is not an unfair additional advantage acquired from the testing domain. The use of mini-batch information, including the feature mean, is a common practice in deep models, as exemplified by techniques like batch normalization. It is interesting that batch normalization eliminates the first-order statistic information from the features to aid in generalization while our method incorporates it as the domain information for the same purpose of enhancing generalization.

\subsubsection{Module Structure Design}
The structure of the adaptation module $g$ is another crucial consideration. One straightforward approach is to concatenate the feature vector and the mini-batch feature mean, and then employ an MLP to generate a new vector with the same dimension as the original feature vector. 
However, this design maps the original feature vectors into a completely different vector space. Since the pre-trained meta-source classifier remains fixed for the purpose of generalizable self-adaptation, the ideal objective is to modify the feature vectors in a controlled manner while keeping them in the same vector space as their original representation. 
Moreover, from the perspective of preserving the prediction ability on the meta-source domain, 
the output vectors from the self-adaptation module should also reside in the same vector space as the input vectors.

For this purpose, we draw inspiration from the concept of feature-wise linear modulation (FiLM) \cite{perez2018film} 
and propose a context-aware feature-wise linear modulation (CaFiLM) method
for our self-adaptation module. 
Specifically, let $\boldsymbol{z}=f_i(x)$ represent the feature vector with dimension size $C$
extracted from the pre-trained meta-source feature extractor $f_i$ for 
the input instance $x$.
Each individual feature dimension, 
denoted as $\{{z}_c\}_{c=1}^C$,
is modulated using a pair of dimension-specific weight 
${\gamma}_c$ and bias ${\beta}_c$, i.e.
\begin{equation}
    CaFiLM({z}_c) = {\gamma}_c {z}_c + {\beta}_c
\end{equation}
where $c$ denotes the dimension index.
Importantly, the weight ${\gamma}_c$ and bias ${\beta}_c$ are learned from the $c$-th dimensional feature 
${z}_c$ and the $c$-th dimensional mini-batch feature mean ${\mu}_c$,
\begin{equation}
    \begin{bmatrix}
        {\gamma}_c \\
        {\beta}_c
    \end{bmatrix} = 
    {A}^{(2\times 2)} 
    \begin{bmatrix}
        {z}_c \\
        {\mu}_c
    \end{bmatrix} +
    \boldsymbol{b}^{(2\times 1)}
\end{equation}
where the mini-batch feature mean is $\boldsymbol{\mu}=\mathbb{E}_{x\in X_b} f_i(x)$. 
Matrix ${A}$ and vector $\boldsymbol{b}$ are learnable parameters
shared across feature dimensions.

For implementation, feature vector $\boldsymbol{z}$ and mini-batch feature mean $\boldsymbol{\mu}$ are firstly stacked into a two-row matrix. 
Secondly, a one-dimensional convolutional layer with parameters of weight matrix ${A}$ and bias vector $\boldsymbol{b}$ is applied along the stacking dimension to produce the weight vector $\boldsymbol{\gamma}$ and bias vector $\boldsymbol{\beta}$. Finally, the feature vector $\boldsymbol{z}$ is linearly modulated through Hadamard product (element-wise product) with $\boldsymbol{\gamma}$ and addition with $\boldsymbol{\beta}$.

Note that the parameters $\boldsymbol{\theta}_g$ of the self-adaptation module $g$ consist solely of the weight matrix 
${A}$ and bias vector $\boldsymbol{b}$, totally only 6 parameters. This simple structure ensures that the self-adaptation module adds minimal extra cost to the classification model. However, when processing large-scale complex datasets, there may be concerns about the ability of the self-adaptation module to handle the increased complexity and diversity of the data. 
We propose to handle this issue by unfreezing the pre-trained meta-source classifier $h_i$ and fine-tuning its parameters during the training of the self-adaptation module $g$.

\subsection{Inference with Ensemble Mechanism}
Once the context-aware self-adaptation module has been trained, we obtain multiple adapted models denoted as 
$\{(h_i\circ g\circ f_i)\}_{i=1}^{|\mathcal{T}|}$,
where $|\mathcal{T}|$ denotes the number of meta-tasks. 
We deploy an ensemble strategy for the inference process to further improve the test performance. 
In particular, 
a simple averaging ensemble
strategy can be applied to the predicted probability vectors generated by the multiple adapted models, i.e.
\begin{equation}
	\hat{y} = \arg\max_c \left[\mathbb{E}_{i\in\{1:|\mathcal{T}|\}} [(h_i\circ g_{\mathcal{C}(x)}\circ f_i(x))]\right]_c
\end{equation}
It's important to note the distinction between the proposed approach and EoA (Ensemble of Averages) \cite{arpit2021ensemble}. While EoA also employs ensemble methods for inference, the key differences lie in the training process and the use of
training data.
In EoA, multiple prediction models are trained using the same training data but with different random seeds and hyper-parameters. The ensemble of these models aims to capture diverse perspectives and variations that arise from the stochasticity of the training process. 
By contrast,
in the proposed approach, the multiple models are trained with different training data from meta-source domains and subsequently adapted using the context-aware self-adaptation module.
The use of models trained from different meta-source and meta-target data provides distinct aspects and viewpoints on the testing data, 
mimicking the benefits of hearing from experts with different levels and areas of knowledge. This diversity in training data enhances the model's understanding and performance on the unseen testing domain. 
It shares similarity with
the concept of random forests \cite{breiman2001random}, except our approach can work with a limited number of ensemble entities. 
Moreover, the context-aware self-adaptation module contributes to the generalization ability of each adapted model. By incorporating the context information and preserving the prediction ability on the meta-source domain, the self-adaptation module further improves the inference performance.
Overall, the combination of training models with different meta-source data and applying the context-aware self-adaptation module enables a comprehensive and robust inference process, leveraging both the diversity of training data and the generalization capabilities of the adaptation module.

\begin{table*}[t]
\centering
\caption{Average accuracy ($\%$) on the testing domain of all tasks for each dataset. Each experiment is conducted 3 times with random seeds and the standard errors is reported with the average values. Results for the comparison methods are cited from the DomainBed and the original papers. The same backbone ResNet-50 is used for all the methods.}
\label{table-main}
\resizebox{\textwidth}{!}{
\begin{tabular}{lccccccccc}
\toprule
\textbf{Algorithm}        & \textbf{PACS}             & \textbf{VLCS}             & \textbf{OfficeHome}       & \textbf{TerraIncognita}   & \textbf{DomainNet}        & \textbf{Avg}              \\
\midrule
ERM \cite{gulrajani2020search}                 & 85.5 $\pm$ 0.2            & 77.5 $\pm$ 0.4            & 66.5 $\pm$ 0.3            & 46.1 $\pm$ 1.8            & 40.9 $\pm$ 0.1            & 63.3                      \\
I-Mixup \cite{xu2020adversarial}               & 84.6 $\pm$ 0.6            & 77.4 $\pm$ 0.6            & 68.1 $\pm$ 0.3            & 47.9 $\pm$ 0.8            & 39.2 $\pm$ 0.1            & 63.4                      \\
MLDG \cite{li2018learning}                     & 84.9 $\pm$ 1.0            & 77.2 $\pm$ 0.4            & 66.8 $\pm$ 0.6            & 47.7 $\pm$ 0.9            & 41.2 $\pm$ 0.1            & 63.6                      \\
SagNet \cite{nam2021reducing}                  & 86.3 $\pm$ 0.2            & 77.8 $\pm$ 0.5            & 68.1 $\pm$ 0.1            & 48.6 $\pm$ 1.0            & 40.3 $\pm$ 0.1            & 64.2                      \\
CORAL \cite{sun2016deep}                       & 86.2 $\pm$ 0.3            & 78.8 $\pm$ 0.6            & 68.7 $\pm$ 0.3            & 47.6 $\pm$ 1.0            & 41.5 $\pm$ 0.1            & 64.5                      \\
mDSDI \cite{bui2021exploiting}                 & 86.2 $\pm$ 0.2            & 79.1 $\pm$ 0.4            & 69.2 $\pm$ 0.4            & 48.1 $\pm$ 1.4            & 42.8 $\pm$ 0.1            & 65.1                      \\
ITL-Net \cite{gao2022loss}                     & 86.4                      & 78.9                      & 69.3                      & 51.0                      & 41.6                      & 65.4                      \\
MIRO \cite{cha2022domain}                      & 85.4 $\pm$ 0.4            & 79.0 $\pm$ 0.0            & 70.5 $\pm$ 0.4            & 50.4 $\pm$ 1.1            & 44.3 $\pm$ 0.2            & 65.9                      \\
SWAD \cite{cha2021swad}                        & 88.1 $\pm$ 0.1            & 79.1 $\pm$ 0.1            & 70.6 $\pm$ 0.2            & 50.0 $\pm$ 0.3            & 46.5 $\pm$ 0.1            & 66.9                      \\
DNA \cite{chu2022dna}                          & 88.4 $\pm$ 0.1            & 79.0 $\pm$ 0.1            & 71.2 $\pm$ 0.1            & 52.2 $\pm$ 0.4            & 47.2 $\pm$ 0.1            & 67.6                      \\
EoA \cite{arpit2021ensemble}                   & 88.6                      & 79.1                      & 72.5                      & 52.3                      & {\bf 47.4}                 & 68.0                       \\
\midrule
Ensemble (Ours)                                & 87.8                      & 80.9                      & 73.0                      & 49.0                      & 44.7                       & 66.8                     \\
CASA (Ours)                                    & {\bf 89.7 $\pm$ 0.1}      & {\bf 81.5 $\pm$ 0.1}     & {\bf 73.5 $\pm$ 0.2}      & 52.0 $\pm$ 0.2            & 47.2 $\pm$ 0.1            & {\bf 68.8}                      \\
\bottomrule
\end{tabular}}
\end{table*}

\section{Experiments}

\subsection{Experimental settings}

Experiments were carried out using the domain generalization test-bed called DomainBed \cite{gulrajani2020search}.

{\bf Datasets.} 
We conducted evaluations of our method on five diverse datasets: PACS, VLCS, OfficeHome, TerraIncognita, and DomainNet. These datasets provide a wide range of domains and classes, allowing us to assess the performance and generalization capabilities of our method across different scenarios.
The PACS dataset \cite{li2017deeper} consists of 9,991 images categorized into 7 classes, spanning four domains: art, cartoons, photos, and sketches.
The VLCS dataset \cite{fang2013unbiased} comprises 10,729 images categorized into 5 classes across four domains: Caltech101, LabelMe, SUN09, and VOC2007.
The Office-Home dataset \cite{venkateswara2017deep} includes 15,588 images across 65 classes and four domains: art, clipart, product, and real.
The TerraIncognita dataset \cite{beery2018recognition} consists of 24,788 images of wild animals, covering 10 classes and captured from four camera locations: L100, L38, L43, and L46.
The DomainNet dataset \cite{peng2019moment} is a large-scale dataset with 586,575 images and 345 classes, encompassing six domains: clipart, infograph, painting, quickdraw, real, and sketch.

{\bf Implementation details.} 
We perform multiple training and single testing domain generalization tasks. 
For each task, we create meta-source and meta-target domains using the training domains. 
Specifically, there are seven combinations of meta-source and meta-target domains for the first four datasets.
For the DomainNet dataset, we select 11 combinations from the 31 possible combinations of training domains. 
We use ResNet-50 as the classification model for all experiments. 
For meta-source model training, models are optimized with the Adam optimizer with learning rate $5\times 10^{-5}$, batch size 32 for the first 4 datasets and 16 for DomainNet dataset. 
For the context-aware self-adaptation module training, we employ the same optimizer and batch size. The parameters of the meta-source classifiers are frozen for the PACS, VLCS, and OfficeHome datasets. For the TerraIncognita and DomainNet datasets, we fine-tune the parameters of the meta-source classifiers when training the self-adaptation module. 
The learning rate for fine-tuning the meta-source classifier parameters is set to $5\times 10^{-5}$, while for the self-adaptation module parameters, it is set to $10^{-3}$.
The trade-off parameter for the preserving loss is set as 0.1 for PACS dataset and 1 for other datasets.
The model selection method here is to use the held-out validation sets from the training data. 

\subsection{Results}

\begin{table*}[t]
\centering
\caption{Average accuracy ($\%$) on the testing domain of all tasks for each dataset. Ablation for adaptation training.}
\label{ablation-adaptation}
\resizebox{0.8\textwidth}{!}{
\begin{tabular}{lccccccccc}
\toprule
\textbf{Algorithm}        & \textbf{PACS}             & \textbf{VLCS}             & \textbf{OfficeHome}       & \textbf{TerraIncognita}   & \textbf{DomainNet}        & \textbf{Avg}              \\
\midrule
Ensemble ($h_i \circ f_i$)                     & 87.8                      & 80.9                      & 73.0                      & 49.0                      & 44.7                       & 66.8                     \\
Ensemble ($h_i\circ g_i\circ f_i$)             & 88.4                      & 80.1                      & 72.5                      & 51.0                      & 43.1            & 67.0                     \\
CASA                                           & {\bf 89.7}                & {\bf 81.5}                & {\bf 73.5}                & {\bf 52.0}                & {\bf 47.2}      & {\bf 68.8}                      \\
\bottomrule
\end{tabular}}
\end{table*}

\begin{table*}[t]
\begin{minipage}[]{0.5\textwidth}
\centering
\caption{Accuracy (\%) on the 4 tasks of OfficeHome dataset.}
\label{ablation-context}
\begin{tabular}{lccccc}
\toprule
\textbf{Algorithm}   & \textbf{A}           & \textbf{C}           & \textbf{P}           & \textbf{R}           & \textbf{Avg}         \\
\midrule
w/o context      & 68.3       & 57.5       & 81.1       & 82.8       & 72.4  \\
w/ context       & {\bf 70.5} & {\bf 58.0} & {\bf 82.0} & {\bf 83.4} & {\bf 73.5}  \\
\bottomrule
\end{tabular}
\end{minipage}
\hfill
\begin{minipage}[]{0.5\textwidth}
\centering
\caption{Accuracy (\%) on the 4 tasks of OfficeHome dataset.}
\label{ablation-film}
\begin{tabular}{lccccc}
\toprule
\textbf{Algorithm}   & \textbf{A}           & \textbf{C}           & \textbf{P}           & \textbf{R}           & \textbf{Avg}         \\
\midrule
MLP     & 68.5       & 57.3       & 81.3       & 82.5       & 72.4  \\
CaFiLM    & {\bf 70.5} & {\bf 58.0} & {\bf 82.0} & {\bf 83.4} & {\bf 73.5}  \\
\bottomrule
\end{tabular}
\end{minipage}
\end{table*}

To assess the effectiveness of the proposed method, domain generalization experiments are conducted on all five datasets. 
The results are reported as the average accuracy of all tasks for each dataset. 
Table~\ref{table-main} presents the results obtained from these experiments.
Firstly, we evaluate the ensemble of the multiple pre-trained meta-source models.
The ensemble model, as shown in the "Ensemble (Ours)" row, demonstrates a notable improvement in performance compared to the baseline ERM approach. On average, our ensemble method achieves a significant gain of 3.5 percentage points (pp) in accuracy.
This highlights the remarkable impact of model ensemble. The diversity of the meta-source models, obtained through training on different combinations of domains, allows them to provide varied perspectives and insights into the unseen testing data. 
By aggregating the predictions of these models, we obtain more accurate decisions than relying solely on a single expert with the most knowledge.
Furthermore, when comparing our simple ensemble approach to previous domain generalization methods, it outperforms all methods prior to MIRO. This further emphasizes the effectiveness of the ensemble model in improving generalization performance in the domain generalization setting.

Secondly, we evaluate the proposed Context-Aware Self-Adaptation method.
Based on the evaluation of the proposed method, the "CASA (Ours)" row in the results table demonstrates a significant performance gain compared to the baseline ERM approach. Our method achieves an impressive performance improvement of 5.5 percentage points (pp) on average. 
This performance gain can be attributed to two key factors: the ensemble of multiple models and the context-aware self-adaptation module. As mentioned earlier, the ensemble approach, as shown in the "Ensemble (Ours)" results, already provides a 3.5 pp improvement compared to the simple ERM baseline.
The additional 2 pp gain is a result of the context-aware self-adaptation module. CASA adapts each pre-trained meta-source model to the meta-target domain while preserving the prediction ability on the meta-source domain. The context information of mini-batch feature mean, treated as domain knowledge, is incorporated into the self-adaptation module to enable self-adaptation. Through exposure to multiple pairs of meta-source models and meta-target domain data, the module learns how to process domain knowledge and modulate the meta-target features, allowing the adapted meta-source model to perform well on both the meta-source and meta-target domains.
In the face of new batches of unseen testing data, the CASA method automatically incorporates context information and aids in the adaptation of the meta-source models, enabling them to generalize well on the testing data.
Compared to previous methods, our proposed CASA method achieves state-of-the-art results on the first three datasets and demonstrates the best average accuracy overall.

\subsection{Ablation study}

\paragraph{Is adaptation training necessary?} 
To rigorously evaluate the necessity of self-adaptation module training, an ablation study is conducted. 
In this study, the self-adaptation module is added as an component of each meta-source feature extractor during meta-source model training, and the ensemble of the meta-source models 
is evaluated. 
This analysis aims to assess the impact of incorporating the self-adaptation module at different stages and determine its necessity in improving the overall performance of the ensemble model.
The results are shown in Table~\ref{ablation-adaptation}.
Upon integrating the adaptation module into the meta-source feature extractor, the ensemble model produces comparable results to the ensemble model without the adaptation module. This indicates that the module alone does not enhance the model's generalization ability. However, through the proposed CASA approach, the self-adaptation module is trained using multiple pairs of meta-source model and meta-target data. The results demonstrate that training the module with adaptation improves the model's generalization to unseen domains.

\paragraph{Is context information necessary?} 
Incorporating the context information of the mini-batch feature mean into the self-adaptation module serves as domain information and enhances the generalization ability. To determine the necessity of this context information, experiments are conducted by removing it from the self-adaptation module. Since only feature vectors are provided as input to the module, an MLP network is used as a replacement for the FiLM structure to adapt the meta-source models to the meta-target domains. The results of these experiments are presented in Table~\ref{ablation-context}.
The results demonstrate that self-adaptation training with context information significantly improves the performance across all the tasks. 
This indicates the importance and effectiveness of incorporating the context information into the self-adaptation module. 
The improved performance suggests that the module is able to leverage the domain-specific information provided by the mini-batch feature mean to enhance the model's generalization ability and achieve better results on various tasks.

\paragraph{Self-adaptation module structure: CaFiLM or MLP?} 
To examine the impact of the model structure in the self-adaptation module, we perform experiments where an MLP is employed as a replacement for the proposed CaFiLM structure utilized in our method. The ensemble models are then evaluated using this modified setup. The results of these experiments are presented in Table~\ref{ablation-film}.
The results indicate that replacing the CaFiLM structure with an MLP leads to a decrease in performance. 
As previously described, CaFiLM is designed to learn channel-wise modulation weights and biases. 
It linearly modulates meta-target features, ensuring that the modulated features remain in the same vector space as the features prior to adaptation. This process facilitates effective feature adaptation, contributing to improved performance. In contrast, using an MLP in place of CaFiLM lacks the same capability to preserve the vector space and may result in suboptimal performance.

\section{Conclusion}

In this paper, we propose a novel two-stage approach for domain generalization called Context-Aware Self-Adaptation (CASA). 
The proposed method simulates a meta-generalization scenario and incorporates a self-adaptation module to adjust pre-trained meta-source models to meta-target domains while preserving their performance on the meta-source domains. 
The self-adaptation process leverages contextual information, such as mini-batch feature means, as domain knowledge to automatically adapt the model to new contexts. 
To ensure the adapted meta-target feature vectors remain in the same vector space as the original vectors, we propose a 
feature-wise linear modulation mechanism to linearly modulate the meta-target features.
Finally, we employ an ensemble of multiple meta-source models for inference on the testing domain. 
Experimental results validate the effectiveness of our approach, achieving state-of-the-art performance on standard benchmarks.

\bibliography{mybib}
\bibliographystyle{icml2023}

\newpage
\appendix
\onecolumn

\end{document}